%% file: main.tex
\documentclass[10pt,twocolumn,letterpaper]{article}

\usepackage[iccvfinal]{iccv}      % To produce the REVIEW version

\definecolor{iccvblue}{rgb}{0.21,0.49,0.74}
\usepackage[pagebackref,breaklinks,colorlinks,allcolors=iccvblue]{hyperref}
\usepackage{multirow}
\usepackage{array}
\usepackage{hyperref}
\usepackage{utfsym}

\title{FastMap: Fast Queries Initialization Based Vectorized HD Map Reconstruction Framework}

\author{Haotian Hu\\
{\tt\small hu\_haotian@leapmotor.com}
\and
Jingwei Xu\\
{\tt\small xu\_jingwei@leapmotor.com}
\and
Fanyi Wang\\
{\tt\small 11730038@zju.edu.cn}
\and
Toyota Li\\
{\tt\small independent}
\and
Yaonong Wang\\
{\tt\small wang\_yaonong@leapmotor.com}
\and
Laifeng Hu\\
{\tt\small hu\_laifeng@leapmotor.com}
\and
Zhiwang Zhang\\
{\tt\small zhiwang.zhang@nit.zju.edu.cn}
}

\begin{document}
\maketitle
\begin{abstract}

Reconstruction of high-definition maps is a crucial task in perceiving the autonomous driving environment, as its accuracy directly impacts the reliability of prediction and planning capabilities in downstream modules. Current vectorized map reconstruction methods based on the DETR framework encounter limitations due to the redundancy in the decoder structure, necessitating the stacking of six decoder layers to maintain performance, which significantly hampers computational efficiency. To tackle this issue, we introduce \textbf{FastMap}, an innovative framework designed to reduce decoder redundancy in existing approaches. FastMap optimizes the decoder architecture by employing a single-layer, two-stage transformer that achieves multilevel representation capabilities. Our framework eliminates the conventional practice of randomly initializing queries and instead incorporates a heatmap-guided query generation module during the decoding phase, which effectively maps image features into structured query vectors using learnable positional encoding. Additionally, we propose a geometry-constrained point-to-line loss mechanism for FastMap, which adeptly addresses the challenge of distinguishing highly homogeneous features that often arise in traditional point-to-point loss computations. Extensive experiments demonstrate that FastMap achieves state-of-the-art performance in both nuScenes and Argoverse2 datasets, with its decoder operating \textbf{$\times$} \textbf{3.2 faster} than the baseline. Code and more demos are available at \href{https://github.com/hht1996ok/FastMap}{https://github.com/hht1996ok/FastMap}.
\end{abstract}

\section{Introduction}
\label{sec:intro}
%高精地图作为自动驾驶系统的环境先验知识库，通过精准的道路拓扑建模与交通规则编码，为车辆定位、轨迹预测及运动规划等关键模块提供基础支撑。传统基于SLAM的离线建图方案[3]虽能构建厘米级精度地图，但其依赖专业采集设备与人工后处理的特性导致成本居高不下，难以满足自动驾驶量产需求。近年来，基于车载环视相机的在线高精地图重建技术[4,5]通过视觉BEV（Bird's Eye View）表征学习，实现了动态环境下的实时矢量化地图生成，其中MapTR[6]及其衍生工作[7,8]建立的DETR式解码范式成为主流方案。
The high definition map serves as an environmental prior knowledge base for autonomous driving systems, providing foundational support for key modules such as vehicle localization, trajectory prediction, and motion planning through precise road topology modeling and traffic rule encoding. Traditional off-line mapping solutions based on SLAM~\cite{shan2018slam2, shan2020slam1}, while capable of building centimeter-level accuracy maps, rely on specialist collection equipment and manual post-processing, resulting in high costs that make it difficult to meet the mass production demands of autonomous driving. In recent years, significant progress has been made in the field of online high-definition map reconstruction technologies based on in-vehicle surround-view cameras~\cite{li2022hdmapnet,liao2024maptrv2,chen2024maptracker,hu2024admap,wu2024lgmap,li2024mapnext,wang2024stream,li2024dtclmapper}. These technologies have achieved real-time vectorized map generation in dynamic environments through visual BEV (Bird's Eye View) representation learning. Among these, the DETR-style~\cite{carion2020detr, yao2021efficientdetr, zhu2020deformable} decoding paradigm established by MapTR~\cite{maptr} and its derivative works~\cite{liao2024maptrv2,hu2024admap,BeMapNet} have become the mainstream solution.
\begin{figure}[t]
\begin{center}
\includegraphics[width=\linewidth]{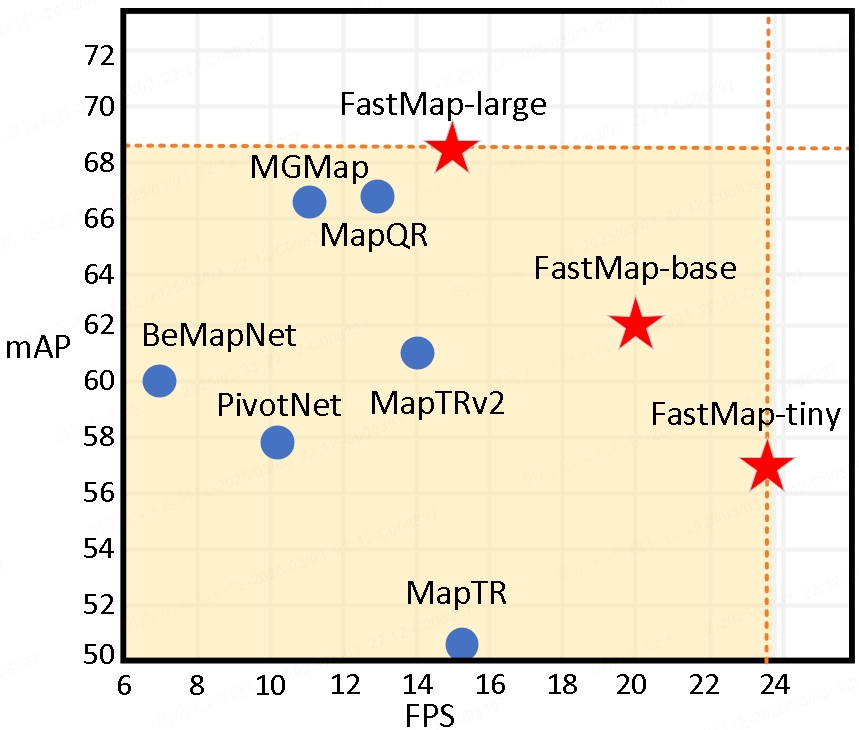}
\end{center}
   \caption{Comparison graphs of mAP and FPS between FastMap and existing models.} 
   \label{fig:1}
\end{figure}

%尽管取得显著进展，现有方法仍面临两大核心挑战：解码器效率瓶颈与预测点被强行拟合到标注点问题。具体而言，MapTR等方法[]为保障地图元素（车道线、路沿等）的预测精度，需堆叠6层Transformer解码器，通过逐层优化随机初始化的查询向量（Query）来渐进式修正预测结果。这一设计虽能提升特征表达能力，却导致模型参数量与计算量很大，严重制约部署效率。受限于车载芯片算力，车端模型通常只能使用一层或两层decoder，这导致其精度显著下降。近期TransFusion等模型[11]在3D检测任务中证实，通过特征引导的查询初始化策略，单层解码器即可达到多层级联结构的性能，这为高效解码器设计提供了新思路。但其在矢量化地图重建中的适用性尚未得到验证，并且需要适合的查询初始化方法来适配地图实例的特殊形状。另一方面，现有监督范式强制模型拟合人工标注的离散矢量点坐标，然而地图元素上矢量点之间的并不存在显著的语义差异。此外，强行将预测点拟合到真值点过于严格的约束了模型的输出，这不利于模型的学习过程。
Despite significant progress that has been made, existing methods still face two core challenges: the efficiency bottleneck of the decoder and the issue of forcibly fitting predicted points to annotated points. Specifically, methods such as MapTR~\cite{maptr} require the stacking of six decoder layers to ensure the accuracy of the prediction of the map elements. This is achieved by progressively refining the randomly initialized query vectors through layer-by-layer optimization. Although this design improves the representation capabilities of the model, it results in a substantial increase in the number of model parameters and the computational load, severely constraining deployment efficiency. Due to computational constraints of onboard vehicle chips, in-vehicle models are typically limited to use only one or two decoder layers, which results in substantial degradation in their accuracy. Recent methods, such as TransFusion~\cite{bai2022transfusion}, have demonstrated in 3D detection tasks that a single-layer decoder can achieve performance comparable to multilayer cascaded structures through feature-guided query initialization strategies, providing new insights for efficient decoder design. However, its applicability in vectorized map reconstruction remains to be validated, and suitable query initialization methods are required to adapt to the unique shapes of map instances. Furthermore, despite the obvious lack of semantic differences between vector points on map elements, the existing supervised paradigm forces the model to adapt to manually annotated vector point coordinates. Forcing the alignment of predicted points with ground-truth points would impose too strict constraints on the model's output, to the detriment of the learning process.

%为此，本文创新性的提出了高效的矢量地图重建框架FastMap。FastMap的解码器仅使用单层双阶段结构的transformer，我们通过heatmap-guided query generation module实现了跨模态注意力机制实现查询向量的智能初始化。具体来说，FastMap解码器在第一阶段通过BEV特征交叉注意力生成具有几何先验的初始化地图查询，取代传统随机初始化方式；第二阶段利用地图查询与图像BEV特征间的位置信息作为先验，直接在ROI区域中生成参考点以进一步细化和完善地图元素特征。在监督层面，我们提出基于点到线的FastMap loss，通过最小化预测点到真值曲线的距离建立监督信号，有效缓解了的预测点被强行拟合到标注点所造成的影响。
To address these challenges, this paper innovatively proposes an efficient vector map reconstruction framework called FastMap. The FastMap decoder uses only a single-layer, two-stage structure of the transformer, and we implement a cross-modal attention mechanism for intelligent initialization of query vectors through the heatmap-guided query generation module. Specifically, the FastMap decoder generates initialized map queries with geometric prior through BEV feature cross-attention in the first stage, replacing the traditional random initialization approach. In the second stage, it leverages the positional information between map queries and image BEV features as a prior to directly generate reference points within the ROI for further refinement and enhancement of map element features. At the supervision level, we propose the FastMap loss based on point-to-line distance, establishing supervision signals by minimizing the distance from predicted points to ground truth curves. This effectively mitigates the adverse effects of forcibly fitting predicted points to annotated points.

%我们在nuScenes和Argoverse2数据集中验证了FastMap的高效性和有效性。如图1所示，FastMap-large实现了现有模型中的最佳性能，其在nuScenes基准中达到了（）mAP。Fastmap-base全面超越了MapTRv2，在提高mAP的情况下使推理速度提高40.3%。
We validated the efficiency and effectiveness of FastMap in the nuScenes~\cite{caesar2020nuscenes} and Argoverse2~\cite{wilson2023argoverse} datasets. As shown in Figure \ref{fig:1}, FastMap-large achieved the best performance compared to sota methods, reaching 68.1 mAP in the nuScenes benchmark. Furthermore, FastMap-base exhibited substantial superiority over MapTRv2, enhancing mAP while concomitantly increasing the inference speed by 40.3\%. Our contributions are summarized as follows:
\begin{itemize}
    \item We propose a framework that employs heatmap to rapidly initialize map queries, which enables our method to achieve better performance using a single layer decoder than the multilayer decoder baselines.
    
    \item We construct a point-to-line based loss mechanism, which is different from the traditional point-by-point coordinate regression based method whose accuracy is largely affected by the accuracy of heatmap. Our method constrains the spatial positions by strictly matching the predicted and labeled points, and transforms the optimization objective into minimizing the distance from the predicted points to the target curve.
    
    \item FastMap significantly improves model inference speed while maintaining model accuracy. Especially for in-vehicle applications, it can greatly enhance online map quality with a negligible increase in time consumption.
\end{itemize}

\section{Related Work}
\label{sec:relate}
\subsection{BEV Perception Decoder}
%BEV感知中的解码器作为BEV感知模型的核心组件，其设计直接影响目标几何与语义信息的恢复精度。DETR [9] 率先将Transformer架构引入2D检测任务，通过多层解码器迭代优化可学习查询向量（Query），但其收敛速度慢、计算冗余的问题在3D场景中愈发显著。PETR [15] 提出3D位置感知解码机制，通过可学习锚点初始化查询并与3D空间特征交互，但仍需6层以上解码器堆叠。针对效率瓶颈，Efficient DETR【】通过滑动窗口在图像特征中提取出密集区域候选框，并根据置信度分数选取候选框作为初始对象查询，显著减少了模型decoder层数。TransFusion [11] 在3D目标检测中验证了特征引导的查询初始化策略，仅用单层解码器即可实现与多层结构相当的精度，这为高效解码器设计提供了新方向，但其在矢量化地图重建中的潜力尚未被探索。
In the domain of BEV perception, the decoder serves as a key component, exerting a direct influence on the accuracy of the target geometry and the recovery of semantic information. DETR~\cite{carion2020detr} pioneered the integration of the Transformer architecture into 2D detection tasks, employing multilayer decoders to iteratively refine the learnable query. Nevertheless, its sluggish convergence and computational inefficiency are exacerbated in 3D contexts. To mitigate the efficiency bottleneck, Efficient DETR~\cite{yao2021efficientdetr} leverages sliding windows to extract dense region proposals from image features and selects candidate boxes as initial object queries based on confidence scores, thus substantially reducing the required number of decoder layers. TransFusion~\cite{bai2022transfusion} demonstrated the efficacy of a feature-guided query initialization strategy in 3D object detection, achieving an accuracy comparable to multilayer structures with a single-layer decoder, thus charting a novel trajectory for efficient decoder design. However, its applicability and potential in the reconstruction of vectorized maps remain unexplored.
\subsection{Online High Definition Map Reconstruction}
%随着在线建图需求的增长，端到端矢量化地图重建成为研究热点。VectorMapNet [16] 首次实现从多视图图像到矢量地图元素的直接回归，采用自回归解码器顺序预测多边形顶点，但其对点序列方向敏感且推理速度受限。MapTR [6] 通过对称排列等价建模解决了点序歧义问题，并引入DETR式匹配机制提升收敛效率，但受限于6层解码器的计算负担。HIMap [17] 设计混合查询（HIQuery）联合建模元素级与点级特征，通过交互式注意力增强局部细节；HRMapNet [18] 引入历史栅格地图先验，利用时序信息提升重建一致性；ADMap [19] 提出矢量方向差损失（VDDL），有效抑制点序抖动现象；MapTracker [20] 将建图任务转化为动态跟踪问题，通过记忆潜在编码（Memory Latents）保持跨帧预测稳定性。Mask2Map[21] 通过掩码感知查询生成（IMPNet）与几何特征驱动预测（MMPNet）的分级建模，结合位置查询增强、实例几何特征提取及跨网络去噪训练策略，显著提升模型精度。
With the increasing demand for online mapping, end-to-end vectorized map reconstruction has emerged as a prominent research focus. VectorMapNet~\cite{liu2023vectormapnet} pioneered direct regression from multiview images to vector map elements, employing an autoregressive decoder to sequentially predict polygon vertices. However, this approach exhibits sensitivity to the direction of point sequences and suffers from limited inference speed. MapTR~\cite{maptr} mitigated the ambiguity of point order through symmetric permutation equivalence modeling and incorporated a DETR-style matching mechanism to enhance convergence efficiency, yet it remains constrained by the computational overhead of a six-layer decoder. HIMap~\cite{zhou2024himap} introduced hybrid queries to jointly model element-level and point-level features, augmenting local details through interactive attention mechanisms. HRMapNet~\cite{zhang2024enhancing} integrated historical raster map priors, utilizing temporal information to improve reconstruction consistency. 

\begin{figure*}[t]
\begin{center}
\includegraphics[width=\linewidth]{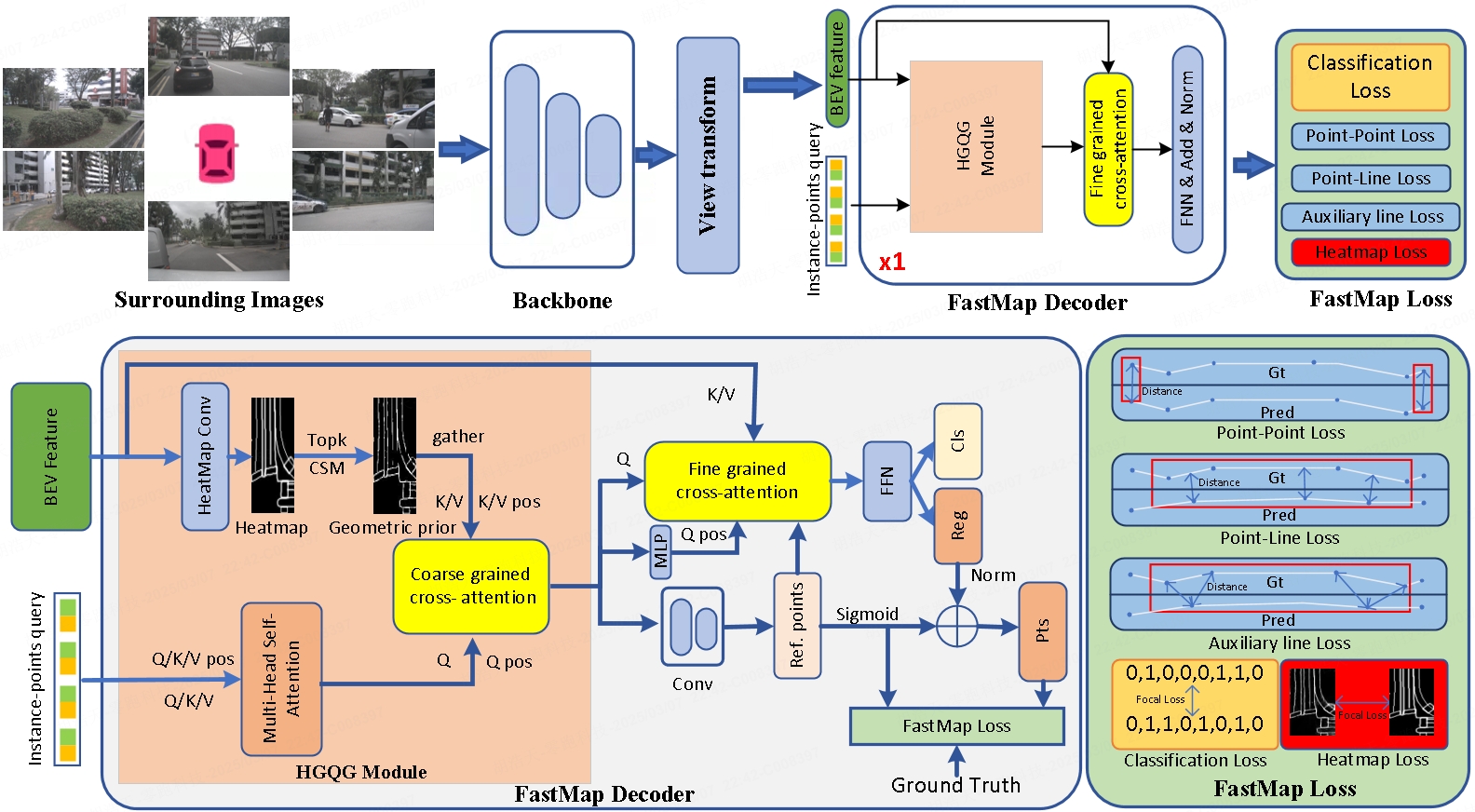}
\end{center}
   \caption{Overall Framework of FastMap. The Decoder Position illustrates how positional information is propagated in the decoder for queries, keys, and values. TopK represents the operation of identifying the Top-K elements with the largest values. CSM denotes the Circular Sampling Method. HGQG Module denotes heatmap-guided query generation module. Gather refers to the operation of extracting elements based on indices. Norm indicates the normalization operation performed according to spatial length.} 
   \label{fig:2}
\end{figure*}

\section{Method}
\label{sec:method}
\subsection{Overall framework}
%如图2所示，FastMap构建了从多相机输入到道路元素感知的端到端框架。骨干网路被用于提取图像特征，视图转换器被用于将多视图特征转换到BEV空间以得到BEV 特征F_BEV。FastMap decoder利用双阶段架构由粗到细的实现查询和BEV特征的交互并生成地图实例。FastMap loss被用来约束模型预测，解决了真值点和语义特征之间模糊关联的问题。
The proposed FastMap is shown in Figure \ref{fig:2}, which establishes an end-to-end framework from multi-camera inputs to road element sensing. Backbone is used to extract image features and view transformer is used to transform multi-view features into BEV space to get BEV feature $F_{BEV}$. The FastMap decoder employs a two-stage architecture that progressively refines the interaction between the query and the BEV features, ultimately generating map instances. FastMap loss mechanism is implemented to constrain the model predictions, thus addressing the challenge of ambiguous associations between ground truth points and semantic features.
\subsection{FastMap Decoder}
%遵循MapTR的设定，我们随机初始化了n*m个Instance-points query以表示地图实例和实例内部点，其中n表示地图实例数量，m表示每个instance内预测点的数量。query和query position表示为Q_feat^init和〖Q_pos^init〗_。为了初始化query可以拥有丰富的语义信息和位置信息，FastMap构建了一个双阶段地图实例解码器。解码器第一阶段利用heatmap-guided query generation module对ROI特征提取策略对鸟瞰图（BEV）特征进行区域聚焦，并有效利用稀疏特征生成粗粒度地图嵌入F_coarse^。该模块中包含了粗粒度交叉注意力（Coarse Grained Cross-Attention, CGCA）和 Geometric Prior Generation策略。输出的嵌入特征包含实例的语义信息和位置信息，为第二阶段多尺度可变形注意力所需的参考点坐标提供支持。
%在解码器第二阶段，我们设计细粒度交叉注意力模块（Fine Grained Cross-Attention, FGCA）实现粗粒度特征与完整BEV特征的深度交互。该模块通过特征补偿机制有效缓解ROI区域提取可能引发的细节丢失问题，最终输出具有完备几何细节和语义信息的地图实例特征F_fine^ 。这种由粗到细的双阶段解码架构在保证计算效率的同时，显著提升了复杂道路拓扑结构的建模能力。

Following the setting of MapTR~\cite{maptr}, we randomly initialized the query $n*m$ instance points to represent map instances and intra-instance points, where $n$ denotes the number of map instances and $m$ denotes the number of predicted points within each instance. The queries and query positions are represented as $Q_{feat}^{init}$ and $Q_{pos}^{init}$, respectively. To ensure that initialized queries are endowed with rich semantic and positional information, FastMap implements a two-stage map instance decoder. 
%In the first stage of the decoder, the self-attention mechanism is employed to capture long-range dependencies among query vectors. Currently, Coarse Grained Cross-Attention (CGCA) is introduced, which utilizes an ROI feature extraction strategy to focus on the foreground within the BEV features, effectively utilizing the sparse features to generate coarse-grained map embeddings $F_{coarse}$. The embedding features produced from this stage contain both semantic and positional information of the instances, supporting the reference point coordinates required for the second phase of multiscale deformable attention.
In the first stage of the decoder, the heatmap-guided query generation module is used to regionally focus the BEV features using the ROI feature extraction strategy and to generate the coarse-grained map embedding $F_{coarse}$ using the sparse features efficiently. The Coarse-Grained Cross-Attention (CGCA) and geometric prior generation strategy are included in this module. The output embedded features contain semantic and location information of the instances, providing support for the reference point coordinates required for the second stage of multi-scale deformable attention.

In the second stage of the decoder, we designed a Fine Grained Cross-Attention (FGCA) to facilitate deep interaction between the coarse-grained features and the complete BEV features. This module effectively mitigates the potential loss of detail caused by ROI extraction through a feature compensation mechanism, ultimately producing map instance features $F_{fine}$ with complete geometric details and semantic information. This two-stage coarse-to-fine decoding architecture significantly enhances the modeling capability of complex road topologies while maintaining computational efficiency.

\begin{figure}[t]
\begin{center}
\includegraphics[width=\linewidth]{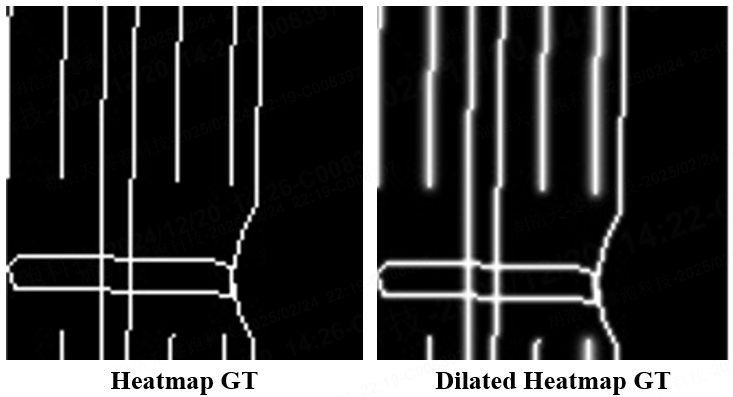}
\end{center}
   \caption{Heatmap ground truth visualisation before and after inflation.} 
   \label{fig:3}
\end{figure}

\subsubsection{Heatmap GT Generation}
%如图3所示，为精准提取鸟瞰图特征中的实例关注区域（ROI），我们针对不同地图元素类别构建多类别感知的实例热力图真值。该真值通过离散化标注点及其拓扑连接生成，我们将实例真值点所在像素及其相邻点连接线覆盖区域设为正样本，同时采用高斯核实现像素膨胀，通过调节核半径动态平衡正负样本比例。
As shown in Figure \ref{fig:3}, to accurately extract ROI from the BEV features, we develop multicategory-aware instance heatmap ground truths specifically designed for different categories of map elements. These ground truths are generated by discretizing annotated points and their topological connections.  We identify the pixels where the ground truth points are located and the regions encompassed by the connecting lines of adjacent points as positive samples. Furthermore, we utilize a Gaussian kernel to achieve pixel expansion, dynamically balancing the ratio of positive to negative samples by adjusting the kernel radius. 

\subsubsection{Geometric Prior Generation}
%BEV特征经过多层卷积网络处理后得到heatmap，其依据置信度阈值筛选出候选ROI区域，同步输出对应位置的采样点特征（sample points features）F_sam及其空间坐标（sample points position）P_sam。
% The BEV features are processed through a multilayer convlutional network to produce a heatmap, which is subsequently employed to filter candidate ROI regions based on a confidence threshold. Simultaneously, the characteristics of the corresponding geometric prior $F_{sam}$ and their spatial coordinates $P_{sam}$ are outputted.

The BEV features are processed using a multilayer convolutional network to generate a heatmap, which is then utilized to filter candidate ROI regions based on a confidence threshold. Concurrently, the characteristics of the corresponding geometric prior $F_{sam}$ and their spatial coordinates $P_{sam}$ are also outputted.
%针对传统均匀采样导致的远距离实例漏检问题，本文提出环形分区采样策略（Circular Sampling Method, CSM）。如图4所示，该方法将鸟瞰图平面划分为L1-L3三个同心圆环区域，基于圆环边缘与BEV中心的距离关系设计渐进式采样规则：外侧最大圆环（L3）采样比例为M*L3/(L3+L2+L1)，内侧最小圆环（L1）对应M*L1/(L3+L2+L1)，其中M为总采样点数。
To address the problem of missed detection of distant instances caused by traditional uniform sampling, we propose a Circular Sampling Method (CSM). CSM divides the BEV plane into three concentric circular regions, labeled $L_1$ to $L_3$. A progressive sampling rule is designed based on the distance relationship between the edges of the circular rings and the center of the BEV: the sampling ratio for the outermost largest ring ($L_3$) is $M\times \frac{L_3}{L_1+L_2+L_3} $, while the ratio for the innermost smallest ring ($L_1$) is $M\times \frac{L_1}{L_1+L_2+L_3} $, where M represents the total number of sampling points. This approach ensures a more balanced and effective sampling strategy across different regions of the BEV plane.
\subsubsection{Coarse Grained Cross-Attention}
%通常，随机初始化query position生成的参考点会使得Deformable Attention难以捕捉到正确的空间位置，这意味着模型需要更多训练轮次来学习偏移量。因此我们提出了Coarse Grained Cross-Attention（CGCA）。该模块通过关联环形采样策略（CSM）输出的M个稀疏采样点，构建粗粒度地图嵌入F_coarse^ 。这些采样点为query提供了大致的位置信息和语义信息。因此，基于粗粒度嵌入生成的参考点可使注意力机制在初始训练阶段即聚焦于关键区域。The pipeline of the decoder can be simply formulated as:
Typically, random initialization of the query position generates reference points that make it difficult for deformable attention to capture the correct spatial location, which means that the model needs more training rounds to learn the offsets. To address this issue, we propose the CGCA. This module constructs coarse-grained map embeddings $F_{coarse}$ by associating with the M sparse sampling points output by the CSM. These sampling points provide the queries with approximate positional and semantic information. Consequently, the reference points generated on the basis of the coarse-grained embeddings enable the attention mechanism to focus on key regions even during the initial training stages. The pipeline of the decoder can be simply formulated as:
\begin{align}
K_{pos} &= W_pP_{sam}, \\
K_{feat} &= F_{sam} + W_cCls,
\end{align}
%其中K_pos表示输入到cross-attention输出的key position，K_feat表示输入的key features，cls表示heatmap中的类别信息。随后计算粗颗粒度实例特征F_Coarse：
where $K_{pos}$ represents the key positions input to the Cross-Attention, $K_{feat}$ denotes the input key features, and $Cls$ signifies the category information in the heatmap. Subsequently, the coarse-grained instance features $F_{coarse}$ are computed as follows:
\begin{align}
\label{eq:2}
K_{coarse}&=K_{feat}+K_{pos}, \\
F_{coarse}&=CA(Q_{pos}^{init},K_{coarse},K_{coarse}).
\end{align}
%其中CrossAttention()按照query、key和value的顺序输入CrossAttention。
In $CA()$ , the inputs are provided in the order of query, key and value to the Cross-Attention.
\subsubsection{Fine Grained Cross-Attention}
%挑选出的采样点可能无法完全包含道路元素中的细节（如大转角或复杂的十字路口等），因此需要进一步的细化地图特征。Fine grained cross-attention（FGCA）由若干个MLP和一个Multi-Scale Deformable Attention组成。为了其更有效的收敛，粗粒度特征F_Coarse经MLP投影生成细粒度查询坐标及参考点偏移量，同时将富含多模态场景特征的鸟瞰图（BEV）特征作为注意力机制的键（Key）与值（Value）输入。细颗粒度实例特征F_Fine表示为：
The selected sampling points may not fully encompass the details within the road elements, which requires further refinement of the map features. The FGCA consists of several MLPs and a multi-scale deformable attention. To facilitate its more effective convergence, the coarse-grained features $F_{coarse}$ are projected through MLP to generate fine-grained query coordinates and reference point offsets. Concurrently, the BEV features, which are rich in multi-modal scene characteristics, are input as the key and value for the attention mechanism. The fine-grained instance features $F_{fine}$ are represented as follows:
\begin{align}
\label{eq:3}
Q_{pos}&=W_{q}F_{coarse}, \\
R_{ref}&=Sigmoid(W_{ref}F_{coarse}),\\
F_{fine}&=MSDeformAttn(Q_{pos},R_{ref},F_{BEV}).
\end{align}
%其中MSDeformAttn()按照query、reference point、value的顺序输入，F_BEV表示 BEV特征，Q_pos表示由粗粒度特征生成的细粒度查询坐标，R_ref为归一化后的参考点。最终通过MLP得到道路元素的位置R和类别信息C。
In this context, the $MSDeformAttn()$ function takes inputs in the order of query, reference point, and value. Here, $F_{BEV}$ represents the BEV features, $Q_{pos}$ denotes the fine-grained query coordinates generated from the coarse-grained features, $W_q$ represents the MLP weights, and $R_{ref}$ represents the normalized reference points.

\subsection{FastMap Loss}
%FastMap Loss消除了人工标注的矢量点带来的影响，利用更合理的点到线之间的联系来约束模型，其中包含了point-line loss、points-points loss、auxiliary line loss、focal loss和heatmap loss。
FastMap loss eliminates the effects of manually annotated vector points and constrains the model with more appropriate point-to-line connections. As illustrated in the bottom right corner of Figure \ref{fig:2}, it consists of five losses, points-points loss, point-line loss, auxiliary line loss, classification loss, and heatmap loss.

\subsubsection{Points-Points Loss}
%考虑到首尾点代表了整个实例的长度，这有别于线上的其他点。因此本文采用混合监督策略，针对首尾端点引入points-points loss L_pp。通过计算预测实例首尾点到真值实例首尾点之间的距离，模型可以更好地约束地图实例的整体长度。
Considering that the start and end points represent the length of the whole instance, which is different from other points on the line. Therefore, we use a hybrid supervised strategy to introduce points-points loss $L_{pp}$ for the start and end points. By calculating the distance between the start and end points of the predicted instance to the start and end points of the ground truth instance, the model can better constrain the overall length of the map instance. This approach ensures that the model accurately captures the spatial extent of the map instances, enhancing the precision and reliability of the reconstructed map.
\begin{align*}
\label{eq:5}
L_{pp}=\sum_{i=0}^{n-1}dis(p_{i,j},gt_{i,j}),j=0,m-1.  \tag{8}
\end{align*}
\begin{table*}[h]
\centering
\small
\caption{Comparison with state-of-the-art methods on the nuScenes validation set. R18, R50, and Effi-B0 denote ResNet18~\cite{he2016resnet}, ResNet50~\cite{he2016resnet}, and EfficientNet-B0~\cite{koonce2021efficientnet} architectures, respectively. R18, R50 and Effi-B0 are all initialized with ImageNet~\cite{deng2009imagenet}  pre-trained weights. FPS (Frames Per Second) metrics are measured on the same machine equipped with an NVIDIA GeForce RTX 3090.}
\begin{tabular}{c|c|c|c|c|c|c|c|c|c|c|c}
\hline
\label{tab:1}
\multirow{2}{*}{Model} & \multirow{2}{*}{Backbone} & \multirow{2}{*}{Epoch} & AP\_div & AP\_ped & AP\_bou & mAP & AP\_div & AP\_ped & AP\_bou & mAP  & \multirow{2}{*}{FPS} \\
\cline{4-11}
& & & \multicolumn{4}{c|}{[0.2m,0.5m,1.0m]} & \multicolumn{4}{c|}{[0.5m,1.0m,1.5m]} & \\ \hline
HDMapNet~\cite{li2022hdmapnet} & Effi-B0 & 30 & 28.3 & 7.1 & 32.6 & 22.7 & 21.7 & 14.4 & 33.0 & 23.0 & 1.6 \\ 
MapTR~\cite{maptr} & R50 & 24 & 30.7 & 23.2 & 28.2 & 27.3 & 51.5 & 46.3 & 53.1 & 50.3 & 15.1 \\ 
PivotNet~\cite{ding2023pivotnet} & R50 & 24 & 42.9 & 34.8 & 39.3 & 39.0 & 58.8 & 53.8 & 59.6 & 57.4 & 10.4 \\ 
BeMapNet~\cite{BeMapNet} & R50 & 24 & 46.9 & 39.0 & 37.8 & 41.3 & 62.3 & 57.7 & 59.4 & 59.8 & 7.0 \\ 
MapQR~\cite{liu2024leveraging} & R50 & 24 & 49.9 & 38.6 & 41.5 & 43.3 & 68.0 & 63.4 & 67.7 & 66.4 & 12.8 \\ 
HybriMap~\cite{zhang2024hybrimap} & R50 & 24 & 45.7 & 38.1 & 42.1 & 41.9 & 65.9 & 63.0 & 67.2 & 65.4 & 9.9 \\ 
PrevPredMap~\cite{peng2024prevpredmap} & R50 & 24 & - & - & - & - & 66.9 & 64.5 & 67.6 & 66.3 & 13.1 \\ 
MGMap~\cite{liu2024mgmap} & R50 & 110 & 54.5 & 42.1 & 37.4 & 44.7 & 67.6 & 64.4 & 67.7 & 66.5 & 11.1 \\ 
PriorMapNet~\cite{wang2024priormapnet} & R50 & 24 & - & - & - & - & 69.0 & 64.0 & 68.2 & 67.1 & 11.7 \\ \hline
MapTRv2~\cite{liao2024maptrv2} & R18 & 110 & - & - & - & - & 55.1 & 46.9 & 54.9 & 52.3 & 16.3 \\ 
MapTRv2~\cite{liao2024maptrv2} & R50 & 24 & 40.0 & 35.4 & 36.3 & 37.2 & 62.4 & 59.8 & 62.4 & 61.5 & 14.1 \\ 
FastMap-tiny & R18 & 24 & 37.4 & 29.5 & 34.4 & 33.8 & 57.4 & 53.2 & 59.6 & 56.7 & 23.3 \\ 
FastMap-base & R50 & 24 & 41.1 & 35.2 & 38.0 & 38.1 & 62.5 & 59.9 & 63.6 & 62.0 & 19.8 \\ 
FastMap-large & R50 & 24 & 54.5 & 44.6 & 41.6 & 46.8 & 69.1 & 65.5 & 69.7 & 68.1 & 14.8 \\ \hline
\end{tabular}
\label{tab:multicol}
\end{table*}
\begin{table}[h!] 
\small
\centering 
\caption{Comparisons with state-of-the-art methods on Argoverse2 val set. All models listed in the table were trained for 6 epochs.}  % 不同self-attention对模型性能的影响。DSA表示decoupled self attention, IIA表示instance interactive attention。
\begin{tabular}{c|c|ccc|c} \hline
\label{tab:2}
Model & Backb.  & $AP_{div}$ & $AP_{ped}$ & $AP_{bou}$ & mAP  \\ \hline
MapTR~\cite{maptr} & R50  &50.6 & 60.7 & 61.2 & 57.4 \\
MapQR~\cite{liu2024leveraging} & R50  & 71.2 & 65.3 & 67.9 & 68.2\\
HIMap~\cite{zhou2024himap} & R50   & 69.5 & 69.0 & 70.3 & 69.6\\ \hline
MapTRv2~\cite{liao2024maptrv2} & R50  & 72.1 & 62.9 & 67.1 & 67.4 \\
FastMap-tiny & R18  & 67.0 & 60.5 & 63.8 & 63.8 \\
FastMap-base & R50  & 71.9 & 63.8 & 68.2 & 68.0\\
FastMap-large & R50 & 72.5 & 69.3 & 72.1 & 71.3\\ \hline
\end{tabular}
\end{table}
%其中distance()代表了预测点和真值点的L1距离。然而在封闭地图元素（人行横道等）中没有首尾点的概念，因此在计算这类封闭地图元素的损失时，我们不会考虑points-points loss L_pp。
Where $dis()$ represents the L1 distance between the predicted point and the ground truth point. However, in closed map elements (such as pedestrian crossing), there is no concept of start and end points. Therefore, when calculating the loss for such closed map elements, the points-points loss $L_{pp}$ is not considered.

\subsubsection{Points-Line Loss}
%在语义特征相似的地图元素上进行点到点的损失计算是不合适的，在地图实例中的点往往拥有非常相似的特征，强行将预测点拟合到真值点是一种输出冗余，这不利于模型学习。为此，本文提出线约束损失函数（Point-Line Loss），采用点到线段的几何约束替代逐点匹配机制。point-line loss L_pl表示如下：
Points in map instances often exhibit highly similar features. Forcibly fitting predicted points to ground truth points creates output redundancy, which is detrimental to model learning. For this reason, this paper proposes a line-constrained loss function (Point-Line Loss), which employs point-to-line geometric constraints instead of the point-by-point matching mechanism. The Point-Line loss $L_{pl}$ is expressed as follows:
\begin{align*}
\label{eq:4}
L_{pl}&=\sum_{i=0}^{n-1}\sum_{j=1}^{m-2}abs(\\&\frac{(p_{i,j,x}-gt_{i,j-1,x})(gt_{i,j,x}-gt_{i,j-1,x})}{\sqrt{(gt_{i,j,y}-gt_{i,j-1,y})^2+(gt_{i,j,x}-gt_{i,j-1,x})^2} } \\& - \frac{ (p_{i,j,y}-gt_{i,j-1,y})(gt_{i,j,y}-gt_{i,j-1,y})}{\sqrt{(gt_{i,j,y}-gt_{i,j-1,y})^2+(gt_{i,j,x}-gt_{i,j-1,x})^2}} ), \tag{9}
\end{align*}
%其中〖gt〗_(i,j,x)和〖gt〗_(i,j,y)表示真值中第i个实例内第j个点的x和y坐标，〖pred〗_(i,j,x)和〖pred〗_(i,j,y)表示预测中第i个实例内第j个点的x和y坐标，n表示instance数量，m表示每个instance内点的数量。
$gt_{i,j,x}$ and $gt_{i,j,y}$ denote the x and y coordinates of the j-th point within the i-th instance in the ground truth, while $p_{i,j,x}$ and $p_{i,j,y}$ represent the x and y coordinates of the j-th point within the i-th instance in the prediction. The variable n indicates the total number of instances and m specifies the number of points contained within each instance.

\subsubsection{Auxiliary Line Loss}
%auxiliary line loss L_al通过计算当前预测点到其对应矢量的两端点间距离之和，约束预测点落在端点连线区间内，从而避免其偏移至线段延长线。其表达式如下：
The auxiliary line loss $L_{al}$ constrains the predicted point to lie within the interval defined by the endpoints of its corresponding vector by calculating the sum of the distances between the current predicted point and the two endpoints of the vector. This prevents the predicted point from deviating onto the extension of the line segment. Its expression is as follows:
\begin{align*}
\label{eq:6}
L_{al}=\sum_{i=0}^{n-1} \sum_{j=1}^{m-2} dis(p_{i,j}-gt_{i,j})+dis(p_{i,j}-gt_{i,j-1}). \tag{10}
\end{align*}
\subsubsection{Heatmap Loss}
%我们使用focal loss[]帮助模型捕捉BEV特征中的地图信息并生成heatmap。为了更好地关注全局的地图元素，FastMap在计算heatmap loss中增加了Gaussian heatmap weight W_gauss来更好地保留远处信息。
We employ focal loss to assist the model capture map information in the BEV features. To better focus on global map elements, FastMap incorporates a Gaussian heatmap weight $W_{gauss}$ in the calculation of the heatmap loss, which enhances the retention of distant information.
\begin{align*}
\label{eq:7}
W_{gauss}&=(1.0-exp(\frac{x^2+y^2}{2\sigma _x\sigma_y } ))\alpha _{gauss}+1.0, \tag{11}
\end{align*}
%其中α_gauss表示高斯权重稀疏，x、y表示当前像素坐标，σ_x=h/β_gauss 和σ_y=w/β_gauss 代表了其分布情况，当β越大，则方差越小。h和w表示heatmap的高和宽。Gaussian heatmap weight从近到远的逐渐增大，在heatmap中心点的值为1.0，最远处其值为α_gauss+1.0。heatmap loss表达式如下所示：
$\alpha_{gauss}$ represents the Gaussian weight sparsity. x and y denote the coordinates of the current pixel. The terms $\alpha _{x}=\frac{h}{\beta _{gauss}}$  and $\alpha _{y}=\frac{w}{\beta _{gauss}}$  define the standard deviations of the Gaussian distribution, where larger $\beta_{gauss}$ results in smaller variances. h and w correspond to the height and width of the heatmap. The weight of the Gaussian heatmap increases gradually from the center to the periphery: it reaches 1.0 at the center of the heatmap and reaches $\alpha_{gauss} + 1.0$ at the farthest edge. The heatmap loss is expressed as follows:
\begin{align*}
\label{eq:8}
L_{focal}&=\sum_{x=0}^{h}\sum_{y=0}^{w}  (H_{x,y}^{p},H_{x,y}^{gt}), \tag{12}
\end{align*}
where $H_{x,y}^{p}$ denotes the predicted of heatmap pixel value and $H_{x,y}^{gt}$ denotes the ground truth of heatmap pixel value.
\subsubsection{Total Loss}
%第一阶段中的粗粒度特征F_Coarse对于模型的快速收敛至关重要，为了得到更准确的参考点位置，我们同样计算第一阶段中生成的参考点和实例真值之间的loss。整体loss L_total表达式如下所示：
In the first stage, the coarse-grained feature $F_{coarse}$ is crucial for rapid convergence of the model. To obtain more accurate reference point positions, we also calculate the loss between the reference points generated in the first stage and the ground truth instances. The overall loss $L_{total}$ is expressed as follows:
\begin{align*}
\label{eq:7}
L_{total}&=\gamma (\alpha_{cls} L_{cls}^{1}+ \alpha_{pl}L_{pl}^{1} +\alpha_{pp}L_{pp}^{1} +\alpha_{al}L_{al}^{1} ) \\& + \beta (\alpha_{cls} L_{cls}^{2} +\alpha_{pl}L_{pl}^{2} +\alpha_{pp}L_{pp}^{2} +\alpha_{al}L_{al}^{2}) 
 \\ &+\alpha_{heat}L_{heat}, \tag{13}
\end{align*}
%其中γ表示第一阶段整体loss的权重，β表示第二阶段整体loss的权重。α_cls、α_pl、α_pp、α_al和α_heat分别表示focal loss、point-line loss、points-points loss、auxiliary line loss和heatmap loss的权重。
where $\gamma$ denotes the weight of the overall loss in the first stage and $\beta$ represents the weight of the overall loss in the second stage. The weights for classification loss, point-line loss, points-points loss, auxiliary line loss and heatmap loss are denoted by $\alpha_{cls}$, $\alpha_{pl}$, $\alpha_{pp}$, $\alpha_{al}$, and $\alpha_{heat}$ respectively.

%##############和tab5合并

\begin{table*}[h!] 
\small
\centering 
\caption{The mAP, inference time, and parameter count of MapTR~\cite{maptr} and FastMap-tiny. Where decoder=n indicates the use of n layer of decoder. When reporting the inference time, we have additionally detailed the FGCA, CGCA, and CSM within the decoder.}  % 不同self-attention对模型性能的影响。DSA表示decoupled self attention, IIA表示instance interactive attention。
\begin{tabular}{c|c|ccc|c|c} \hline
\label{tab:3}
Model & mAP& FGCA & CGCA  & CSM & Decoder & Parameter \\ \hline
MapTR(decoder=1) & 29.1 & 4.1ms& -  & - & 6.2ms & 40.6M \\
MapTR(decoder=2) & 42.4 & 8.1ms&- &-& 10.3ms & 41.8M  \\
MapTR(decoder=6)  & 50.3 & 24.0ms&- &-& 26.2ms & 46.7M   \\
FastMap-tiny  & 56.7 & 2.5ms & 2.9ms  & 1.5ms & 8.1ms & 41.5M\\ \hline
\end{tabular}
\end{table*}

\section{Experiments}
\label{sec:Experiments}
%在本节中，我们首先在第4.1节介绍了实验设置和实现细节。在4.2节中报告了对比实验结果。在4.3节中我们进行了广泛的消融实验以验证各个模块的有效性。4.4节对模型整体进行了分析，进一步证明了其优越性能。
In this section, we first introduce the experimental setup in Section 4.1. In Section 4.2, we report the comparative experimental results. In Section 4.3 we perform extensive ablation experiments to verify the effectiveness of the individual modules. We further analyze the model to demonstrate its superior performance in Section 4.4.
\subsection{Experimental setup}

\begin{table*}[h!] 
\small
\centering 
\caption{Ablation study report for each module. FGCA represents the Fine-Grained Cross-Attention, CGCA denotes the Coarse-Grained Cross-Attention, CSM stands for the Circular Sampling Method, GHW represents the Gaussian heatmap weight, and HP denotes the heatmap pretrain.} 
\begin{tabular}{c|c|c|c|c|c|ccc|c} \hline
\label{tab:4}
FGCA & CGCA  & FastMap & CSM & GHW & HP & $AP_{div}$ & $AP_{ped}$ & $AP_{bou}$ & mAP  \\ \hline
\checkmark & \usym{2715} &\usym{2715} & \usym{2715} & \usym{2715} & \usym{2715} & 30.2 & 20.8 & 36.3 & 29.1 \\
\checkmark & \checkmark  & \usym{2715} & \usym{2715} & \usym{2715} & \usym{2715} & 57.5 & 43.8 & 55.5 & 52.3 (+23.2) \\
\checkmark & \checkmark  & \checkmark & \usym{2715} & \usym{2715} & \usym{2715} & 59.9 & 49.6 & 59.5 & 56.3 (+4.0)\\
\checkmark & \checkmark  & \checkmark & \checkmark & \usym{2715} & \usym{2715} & 58.2 & 55.1 & 62.6 & 58.6 (+2.3)\\
\checkmark & \checkmark  & \checkmark & \checkmark & \checkmark & \usym{2715} &62.0 & 56.7 & 62.3 & 60.4 (+1.8)\\
\checkmark & \checkmark  & \checkmark & \checkmark & \checkmark & \checkmark & 62.5 & 59.9 & 63.6 & 62.0 (+1.6)\\ \hline
\end{tabular}
\end{table*}
\subsubsection{Implementation Details}
%我们使用Resnet50作为FastMap-base和FastMap-large的图像骨干网络，使用Resnet18作为FastMap-tiny的图像骨干网络。优化器为 AdamW，权重衰减为 0.01。批次大小为 32，在训练中使用8个NVIDIA H100进行训练，并且使用单个NVIDIA GeForce RTX 3090 GPU进行推理验证以保证和其他模型测试FPS时条件相同。我们使用12个epoch的heatmap预训练（只使用Heatmap Loss）和24epoch的常规训练，初始学习率为1×10-3，余弦衰减。FastMap-tiny和FastMap-base中的解码器层数设置为1，FastMap-large的解码器层数设置为6。对于nuScenes 数据集和 Argoverse2 数据集的实例查询、点查询数和输入图像尺寸等参数均遵循【MAPTRV2】的设置。FastMap Loss中的α_cls、α_pl、α_pp、α_al和α_heat分别设置为2.0、2.5、2.5、2.5和0.6。heatmap loss的α_gauss设置为0.8。损失权重a和损失权重b分别设置为0.5，1.0。总采样点数M设置为3500。高斯膨胀核尺寸设置为3。
We use ResNet50~\cite{he2016resnet} as the image backbone network for FastMap-base and FastMap-large, and ResNet18~\cite{he2016resnet} for FastMap-tiny. The optimizer is AdamW with a weight decay of 0.01. The batch size is 32, and the training is carried out using 8 NVIDIA H100 GPUs. For inference validation, a single NVIDIA GeForce RTX 3090 GPU is used to ensure consistent conditions with other models when testing FPS. We employ 6 epochs of heatmap pre-training (using only Heatmap Loss) and 24 epochs of regular training, with an initial learning rate of $8×e^{-4}$ and cosine decay. The number of decoder layers is set to 1 for FastMap-tiny and FastMap-base, and to 6 for FastMap-large. For the nuScenes and Argoverse2 datasets, parameters such as instance queries, point query numbers, and input image sizes follow the settings of MapTRv2~\cite{liao2024maptrv2}. In FastMap Loss, the weights $\alpha_{cls}$, $\alpha_{pl}$, $\alpha_{pp}$, $\alpha_{al}$, and $\alpha_{heat}$ are set to 2.0, 2.5, 2.5, 2.5 and 0.6 respectively. The $\alpha_{gauss}$ for heatmap loss is set to 0.8. Loss weight $\alpha$ and loss weight $\beta$ are set to 0.5 and 1.0, respectively. The total number of sampling points M is set to 3500, and the Gaussian dilation kernel size is set to 3.

\subsubsection{Evaluation Dataset}
%我们在主流数据集nuScenes中对我们的方法进行评估。nuScenes数据集中包含二维城市级全局矢量化地图和 1000 个场景，每个场景持续时间约为 20 秒。关键样本的注释频率为 2Hz。每个样本都有来自 6 个摄像头的 RGB 图像，覆盖自我车辆 360◦ 水平 FOV。为了进一步证明模型的鲁棒性，我们在Argoverse2数据集上进行了实验，该数据集包含 1000 条日志。每个日志提供来自 7 个摄像头的 15 秒 20Hz RGB 图像和日志级三维矢量地图。
We evaluated our method on the mainstream nuScenes dataset~\cite{caesar2020nuscenes}. The nuScenes dataset contains 2D city-level global vectorized maps and 1000 scenes, with each scene lasting approximately 20 seconds. Key samples are annotated at a frequency of 2Hz. Each sample includes RGB images from six cameras, covering a 360-degree horizontal Field Of View (FOV) around the ego vehicle. To further demonstrate the robustness of the model, we performed experiments on the Argoverse2 dataset~\cite{wilson2023argoverse}, which contains 1000 logs. Each log provides 15 seconds of 20Hz RGB images from 7 cameras and log-level 3D vectorized maps.
\subsubsection{Evaluation metrics}
%我们将X 轴的感知范围为 [-15.0m, 15.0m]，Y 轴的感知范围为 [-30.0m, 30.0m]，采用平均精度（AP）来评估地图构建质量。倒角距离用于确定预测值与 GT 是否匹配。我们计算多个倒角距离阈值下的 AP，然后取所有阈值的平均值作为最终的 AP 指标。实验中将阈值分别两组，分别是[0.2,0.5,1.0]和[0.5,1.0,1.5]。
We set the perception range for the X-axis to [-15.0m, 15.0m] and for the Y-axis to [-30.0m, 30.0m], using Average Precision (AP) to evaluate the quality of map construction. The chamfer distance is used to determine whether the predicted values match the ground truth (GT). We calculate the AP at multiple chamfer distance thresholds and then take the average of all thresholds as the final AP metric. In the experiments, the thresholds are divided into two groups: [0.2, 0.5, 1.0] and [0.5, 1.0, 1.5].

\subsection{Comparison experiments}
\subsubsection{nuScenes}

%表1中报告了FastMap-tiny、FastMap-base和FastMap-large在nuScenes中的性能指标。FastMap-tiny在使用小骨干网络(Resnet18)的情况下，其模型精度和推理速度远远效果了其基线（MapTR），其精度相较于MapTR提高了6.4mAP，而推理速度显著提升了54.3%。FastMap-base相较于MapTR在提高了0.5mAP的情况下，推理速度大幅提高了40.4%。
%FastMap-large在推理速度略高于基线的情况下取得了最佳性能，其达到了（）mAP，这表明使用heatmap初始化依然能够保证模型的上限。在更严格的评价标注（阈值为[0.2, 0.5, 1.0]）中同样验证了FastMap的优秀性能，对比基线其AP具有更大的提升，这表明FastMap输出的地图实例更为细致。
Table \ref{tab:1} reports the performance metrics of FastMap-tiny, FastMap-base, and FastMap-large on the nuScenes dataset. FastMap-tiny, which uses a small backbone network (ResNet18), significantly outperforms its baseline (MapTR) in both model accuracy and inference speed. Its precision improved by 6.4 mAP compared to MapTR, while the inference speed increased significantly by 54.3\%. FastMap-base, while improving 0.5 mAP over MapTR, achieved a substantial 40.4\% increase in inference speed.
FastMap-large achieved the best performance with a slightly higher inference speed than the baseline, reaching 68.1 mAP, demonstrating that heatmap initialization can still ensure the model's upper limit. The excellent performance of FastMap is also verified in the more strict evaluation annotation (thresholds of [0.2, 0.5, 1.0]), where its AP has a greater improvement compared to the baseline, indicating that FastMap produces more detailed map instances.
\subsubsection{Argoverse2}
%与此同时，我们还在Argoverse2 数据集上评估了我们的框架。FastMap同样展现出了其卓越性能，FastMap-tiny相较于MapTR提高了6.4mAP，FastMap-base相较于MapTRv2提高了0.6mAP，FastMap-large在Argoverse2 数据集上取得了（）mAP。
Meanwhile, Table \ref{tab:2} shows the results of our framework evaluated on the Argoverse2 dataset. FastMap also demonstrated its outstanding performance here. FastMap-tiny improved by 6.4 mAP compared to MapTR, FastMap-base improved by 0.6 mAP compared to MapTRv2, and FastMap-large achieved 71.3 mAP on the Argoverse2 dataset.

\subsection{Ablation experiments}
%表3报告了MapTR和 FastMap-base的mAP，推理时间和参数量。可以看出，FastMap在mAP，运算速度和模型参数量上全面领先于MapTR，在保持性能提高6.4mAP的前提下，其decoder速度相较于MapTR提高17.1ms。FastMap中的FGCA相较于MapTRv2减少了两个MHA（multi-head attention）因此单层耗时减少了1.6ms。
%表4报告了FastMap loss中各个损失的权重分配消融实验。FastMap的α_pl、α_pp、α_al和α_heat分别取2.5,2.5,2.5和0.6时，距离损失和heatmap损失之间达到平衡。
%表5报告了各模块消融实验。我们使用了decoder层数为1的MapTR作为baseline。增加CGCA后模型精度显著提高到52.3mAP，这表明利用ROI区域初始化地图查询可以有效缓解解码器结构冗余问题。FastMap Loss提高了模型4.0mAP，证明了点到线损失有效缓解了人工标注真值造成的特征模糊问题。增加CSM 和heatmap weight loss后模型提高了2.3mAP和1.8mAP，表明这两个策略有效提高了编码器第一阶段的召回率。

Table \ref{tab:3} reports the mAP, inference time, and parameters for MapTR and FastMap-base. It can be seen that FastMap is fully ahead of MapTR in terms of mAP, computing speed and number of model parameters, and its decoder speed is improved by 17.1 ms compared to MapTR while maintaining the performance improvement of 6.4 mAP. FGCA in FastMap reduces two MHAs compared to MapTRv2 thus reducing the single-layer elapsed time by 1.6 ms.
%Table 4 presents the ablation study on the weight allocation of losses in FastMap Loss. When the weights  are set to 2.5, 2.5, 2.5, and 0.6 respectively, a balance is achieved between the distance loss and the heatmap loss.

\begin{figure}[t]
\begin{center}
\includegraphics[width=\linewidth]{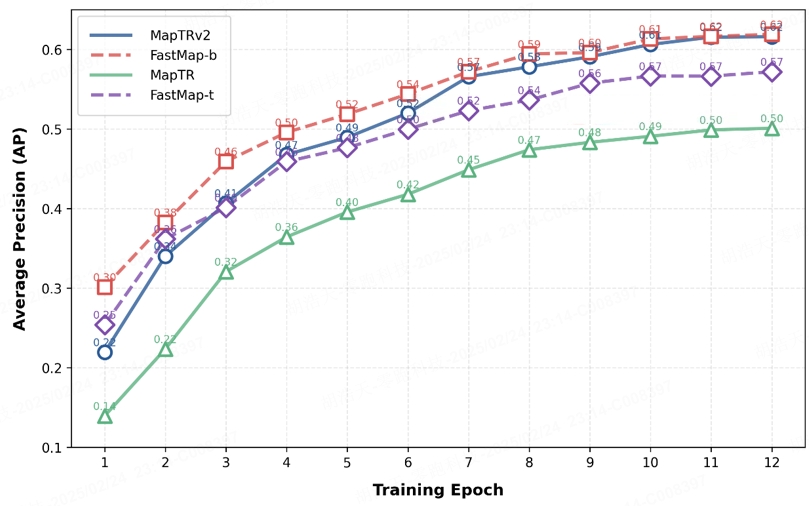}
\end{center}
   \caption{AP rise line graphs for MapTR, MapTRv2, FastMap-tiny, and FastMap-base. x-axis denotes the current epoch and y-axis denotes the corresponding AP.} 
   \label{fig:4}
\end{figure}

Table \ref{tab:4} reports the ablation experiments for each module. We used MapTR with 1 decoder layer as the baseline. Adding CGCA significantly improved the model accuracy to 52.3 mAP, indicating that initializing map queries using ROI regions effectively alleviates the redundancy issue in the decoder structure. FastMap Loss improved the model by 4.0 mAP, proving that the point-to-line loss effectively mitigates the feature ambiguity caused by manual annotation. Adding CSM and heatmap weight loss further improved the model by 2.3 mAP and 1.8 mAP, respectively, demonstrating that these strategies effectively enhance the recall rate in the first stage of the encoder.

\subsection{Analysis}
%图4中可以看出，受益于较为精确的参考点位置，FastMap-base相较于MapTRv2具有更快的收敛速度。在第一个epoch时，两者的mAP相差了8%。同样的，在较为轻量的FastMap-tiny在训练开始阶段的AP甚至超过了MapTRv2，这也表明了FastMap大幅提高了模型收敛速度。
As illustrated in Figure \ref{fig:4}, due to the more accurate location of the reference point, FastMap-base demonstrates a higher convergence rate relative to MapTRv2. In particular, at the first epoch, a disparity of 8.1\% in mAP is observed between the two models. Furthermore, the most lightweight model FastMap-tiny surpasses MapTRv2 in terms of AP during the initial training stage, which underscores the significant improvement in model convergence speed achieved by FastMap.
%表7

\begin{table}[h!] 
\small
\centering 
\caption{Comparison results of ACD/ARD/AJP between FastMap, MapTR and MapTRv2. $\downarrow$ indicates that the lower the value the better the performance.}
\begin{tabular}{c|c|c|c|c} \hline
\small
\label{tab:6}
Model & MapTR & MapTRv2 & FastMap-t & FastMap-b  \\ \hline
ACD $\downarrow$ & 0.597 & 0.560 & 0.551  & 0.505 \\
ARD $\downarrow$ & 7.866 & 5.950 & 6.312 & 5.286 \\
AJP $\downarrow$ & 6.76 & 4.53 & 5.12 & 4.08 \\ \hline
\end{tabular}
\end{table}
ACD refers to the average chamfer distance between each positive sample and the corresponding ground truth instance. ARD measures the average radian distance within an instance, calculated as the sum of the radians of the angles between consecutive line segments in each positive sample and the ground truth segments. AJP represents the average number of jitter points within an instance, defined as points where the angle between adjacent line segments exceeds 30°, while the corresponding true point maintains an angle below 5°. These metrics are systematically compared in Table \ref{tab:6}.
% \subsection{Visualization}
%图6表示预测的heatmap在十字路口中也具有良好的效果，这使得heatmap预测不准不会成为FastMap的瓶颈。此外，CRM均衡的采样了各区域的前景点，这保证了sample points拥有良好的召回率.如图7所示，FastMap所输出的矢量更为平滑并缓解了模型漏检问题。
\begin{figure}[t]
\begin{center}
\includegraphics[width=\linewidth]{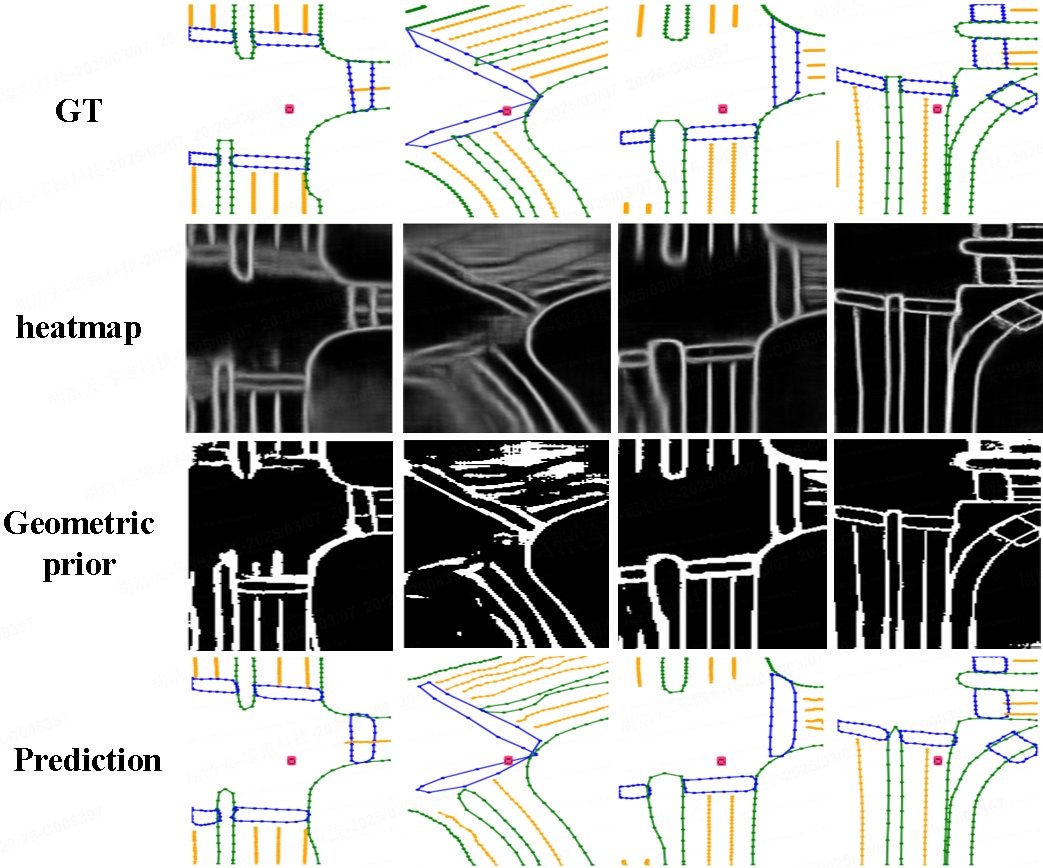}
\end{center}
   \caption{Heatmap ground truth, heatmap, geometric priors and prediction visualization results.} 
   \label{fig:5}
\end{figure}
\begin{figure}[t]
\begin{center}
\includegraphics[width=\linewidth]{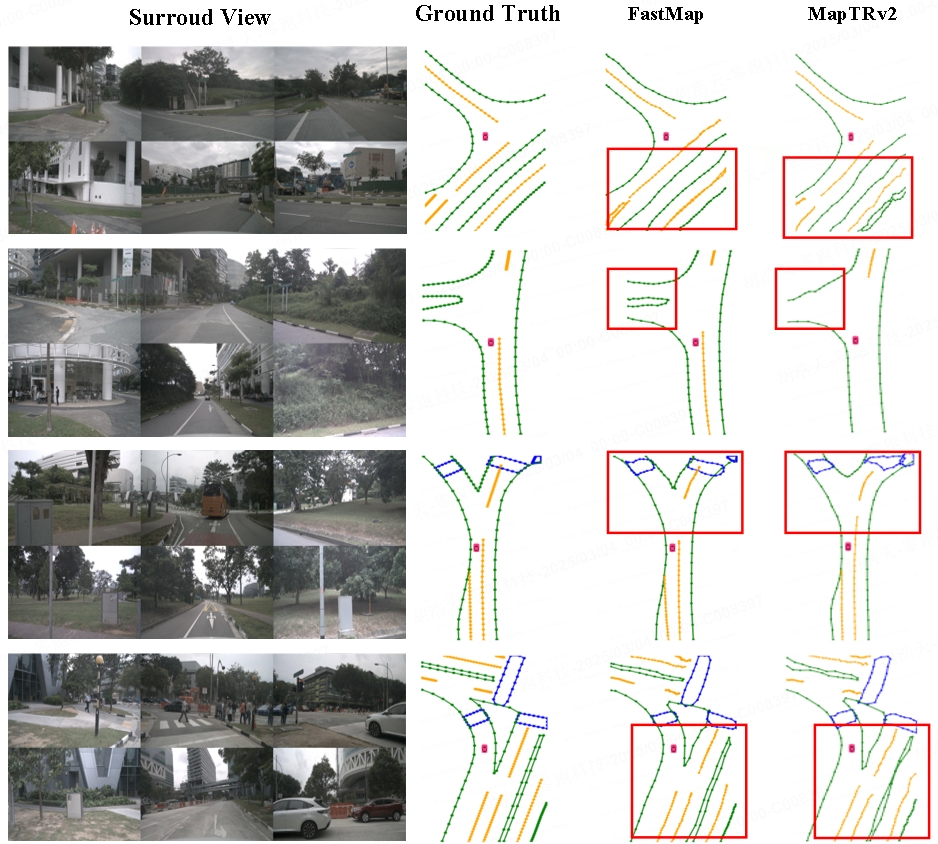}
\end{center}
   \caption{Visualization results of FastMap and MapTRv2 in nuScenes dataset.} 
   \label{fig:6}
\end{figure}

Figure \ref{fig:5} illustrates that the predicted heatmap performs well even in intersections, indicating that inaccurate heatmap prediction does not become a bottleneck for FastMap. In addition, CRM evenly samples foreground prior across all regions, ensuring that the geometric prior maintain a high recall rate. As shown in Figure \ref{fig:6}, the FastMap output vectors are smoother and effectively mitigate the problem of missed detections by the model.
\section{Conclusions}
%FastMap是一种用于高效在线矢量地图的端到端框架，它使用heatmap初始化地图查询以解决解码器冗余堆叠问题并提高模型收敛速度。大量实验证明了我们所提出方法可以在nuScenes和Argoverse2数据集中的实现高效性能。我们希望FastMap可以能为高效在线矢量地图的一种有效范式，能在解决资源有限环境下提供良好性能。并推动下游任务（如运动预测和规划）的发展。 
The Proposed FastMap is an end-to-end framework designed for efficient online vector mapping. It utilizes heatmap initialization for map queries to address the issue of redundant stacking in the decoder and to accelerate the model's convergence speed. Extensive experiments have shown that our proposed method can achieve efficient performance on nuScenes and Argoverse2 datasets. We hope that FastMap can be an effective paradigm for efficient online vector maps that can provide good performance in addressing resource-limited environments. And to advance downstream tasks such as motion prediction and planning.

{
    \small
    \bibliographystyle{ieeenat_fullname}
    \bibliography{main}
}

\end{document}